\documentclass[sigconf]{acmart}

\usepackage{multirow}
\usepackage{setspace}
\usepackage{lipsum}

\AtBeginDocument{%
  \providecommand\BibTeX{{%
    \normalfont B\kern-0.5em{\scshape i\kern-0.25em b}\kern-0.8em\TeX}}}

\newcommand\blfootnote[1]{
\begingroup
\renewcommand\thefootnote{}\footnote{#1}
\addtocounter{footnote}{-1}
\endgroup
}
\copyrightyear{2021}
\acmYear{2021}
\setcopyright{acmcopyright}
\acmConference[MM '21]{Proceedings of the 29th ACM
International Conference on Multimedia}{October 20--24, 2021}{Virtual Event, China}
\acmBooktitle{Proceedings of the 29th ACM International Conference on Multimedia
(MM '21), October 20--24, 2021, Virtual Event, China}
\acmPrice{15.00}
\acmDOI{10.1145/3474085.3475178}
\acmISBN{978-1-4503-8651-7/21/10}

\acmSubmissionID{103}

\begin{document}
\fancyhead{}

\title{Mask is All You Need: Rethinking Mask R-CNN \\ for Dense and Arbitrary-Shaped Scene Text Detection}

\author{Xugong Qin$^{1,2}$, Yu Zhou$^{1,2,*}$, Youhui Guo$^{1,2}$, Dayan Wu$^1$, Zhihong Tian$^3$, Ning Jiang$^4$, \\Hongbin Wang$^4$, and Weiping Wang$^1$
}
\affiliation{
  \institution{
  $^1$Institute of Information Engineering, Chinese Academy of Sciences, Beijing, China\\
  $^2$School of Cyber Security, University of Chinese Academy of Sciences, Beijing, China\\
  $^3$Guangzhou University, Guangzhou, China\\
  $^4$Mashang Consumer Finance Co., Ltd., Beijing, China
  }
  \streetaddress{}
  \city{}
  \state{}
  \country{}
  \postcode{}
}
\email{{qinxugong,zhouyu,guoyouhui,wudayan,wangweiping}@iie.ac.cn}
\email{tianzhihong@gzhu.edu.cn, {ning.jiang02,hongbin.wang02}@msxf.com}








\renewcommand{\shortauthors}{Trovato and Tobin, et al.}

\begin{abstract}
  Due to the large success in object detection and instance segmentation, Mask R-CNN attracts great attention and is widely adopted as a strong baseline for arbitrary-shaped scene text detection and spotting. However, two issues remain to be settled. The first is dense text case, which is easy to be neglected but quite practical. There may exist multiple instances in one proposal, which makes it difficult for the mask head to distinguish different instances and degrades the performance. In this work, we argue that the performance degradation results from the learning confusion issue in the mask head. We propose to use an MLP decoder instead of the ``deconv-conv'' decoder in the mask head, which alleviates the issue and promotes robustness significantly. And we propose instance-aware mask learning in which the mask head learns to predict the shape of the whole instance rather than classify each pixel to text or non-text. With instance-aware mask learning, the mask branch can learn separated and compact masks. The second is that due to large variations in scale and aspect ratio, RPN needs complicated anchor settings, making it hard to maintain and transfer across different datasets. To settle this issue, we propose an adaptive label assignment in which all instances especially those with extreme aspect ratios are guaranteed to be associated with enough anchors. Equipped with these components, the proposed method named MAYOR\footnote{{\bf M}ask is {\bf A}ll {\bf YO}u need: {\bf R}ethinking Mask R-CNN for dense and arbitrary-shaped scene text detection, abbreviated as MAYOR\label{MAYOR}} achieves state-of-the-art performance on five benchmarks including DAST1500, MSRA-TD500, ICDAR2015, CTW1500, and Total-Text.
\end{abstract}

%
%



\maketitle

\begin{small}
\begin{spacing}
1
\textbf{ACM Reference Format:}

\noindent Xugong Qin, Yu Zhou, Youhui Guo, Dayan Wu, Zhihong Tian, Ning Jiang, Hongbin Wang, and Weiping Wang. 2021. Mask is All You Need: Rethinking Mask R-CNN for Dense and Arbitrary-Shaped Scene Text Detection. In \textit{Proceedings of the 29th ACM International Conference on Multimedia (MM ’21), October 20-24, 2021, Virtual Event, China.} ACM, New York, NY, USA, 10 pages. https://doi.org/10.1145/3474085.3475178
\end{spacing}
\end{small}

\blfootnote{* Yu Zhou is the corresponding author.}

\vspace{-10px}
\begin{figure}[!htb]
\begin{center}
    \includegraphics[width=0.90\linewidth]{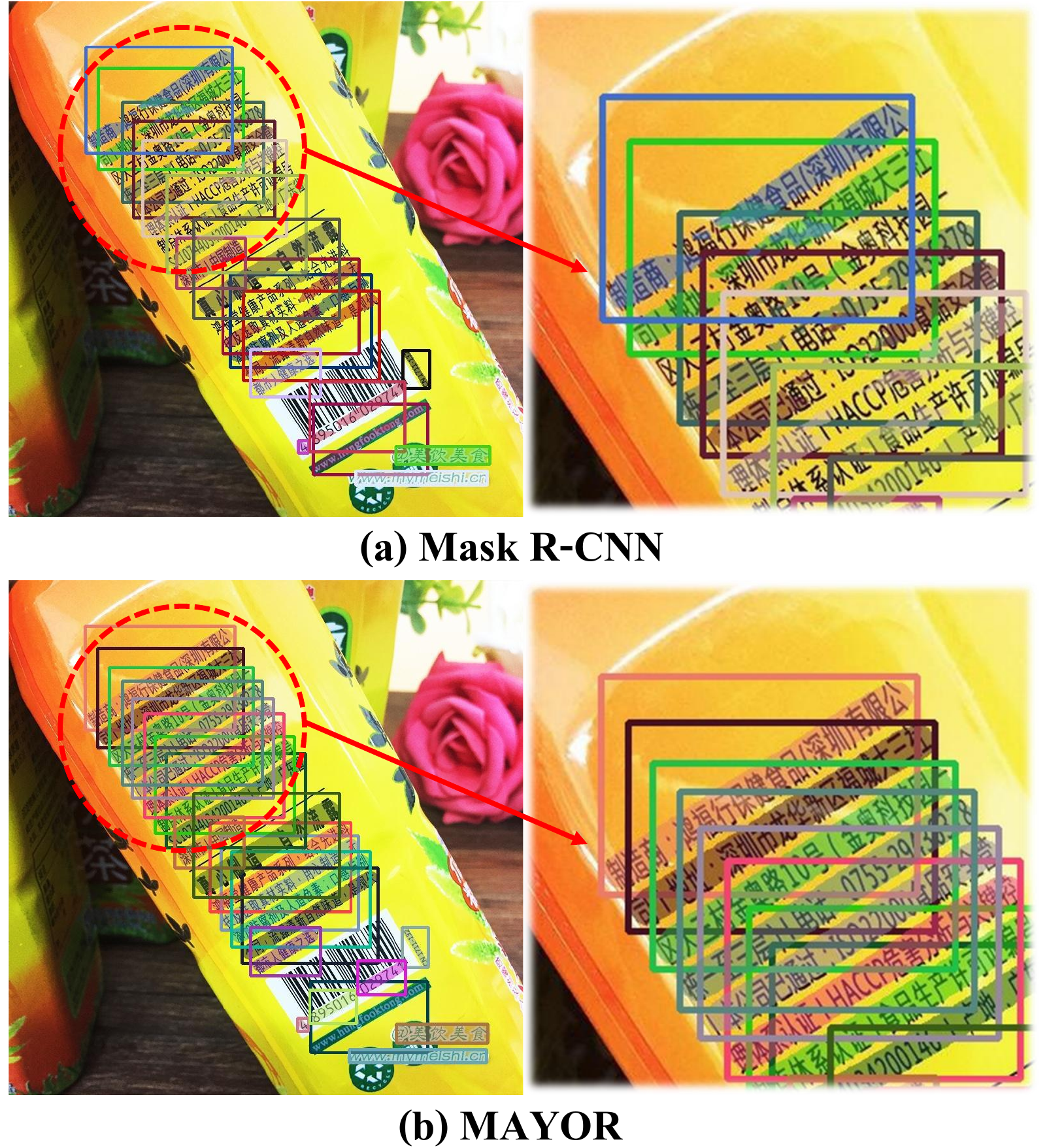}
\end{center}
    \vspace{-10px}
    \caption{Comparison between results produced by different methods on DAST1500. (a) and (b) denote results from Mask R-CNN and MAYOR$^{\ref{MAYOR}}$. Different colors are used to distinguish different instances. The boxes and the masks are the corresponding predicted bounding boxes and masks.}
    \vspace{-15px}
\label{fig:illus}
\end{figure}

\section{Introduction}

Scene text detection (STD) has attracted attention due to its practical applications, e.g., scene understanding, blind navigation, and document analysis. Though great progress has been achieved with CNN-based methods inspired by general object detection and segmentation frameworks like Faster R-CNN \cite{ren2015fasterrcnn}, Mask R-CNN \cite{he2017maskrcnn}, and FCN \cite{long2015FCN}, STD remains challenging due to large variations in scale, orientation, and aspect ratio, as well as arbitrary shape. 

Mask R-CNN, as one of the most powerful detectors for general object detection and instance segmentation, is widely adopted as a strong baseline for arbitrary-shaped scene text detection and spotting \cite{xie2019SPCNet,liu2019BDN,Wang2020contournet,lyu2018masktextspotter,liao2019masktextspotterv2,liao2020masktextspotterv3,qin2019towards,liu2019maskTTD,ye2020textfusenet,liu2020asts}. 
Though excellent performance has been achieved with Mask R-CNN based methods, there are still two challenges to be addressed in dense and arbitrary-shaped scene text detection.


The first challenge is dense text case, which is not received enough attention in previous works \cite{tang2019seglink++,ma2021relatext} and is regarded as one of the key bottlenecks for Mask R-CNN based scene text detection frameworks. Mask R-CNN can not well handle dense text case which may result in multiple instances falling in one proposal as shown in Fig. \ref{fig:illus} (a). Liao et al. \cite{liao2020masktextspotterv3} propose to integrate segmentation-based methods \cite{wang2019PSENet,wang2019PAN,liao2020db} to generate proposals in a bottom-up manner, which can avoid this issue and achieve robust text reading. However, the segmentation proposal network requires sequentially finding connected regions, slowing down the training and inference process. It also suffers from the accumulated errors from the text kernel localization. Different from it, we handle this problem from a top-down perspective. As shown in Fig. \ref{fig:illus}, our motivation comes from an observation: though a box contains multiple text instances, the box branch can localize texts accurately with an axis-aligned rectangle bounding box. Since the RoI feature representation is able to decode the information of the instance in the box branch, it may have considerable representation to recover the whole instance in the mask branch. Thus it is reasonable to delve into the mask branch. We revisit the mask learning process and find the performance degradation is caused by the learning confusion issue in the training process.

The second challenge is that the manually pre-designed anchors cannot easily match text instances of extreme aspect ratios. TextBoxes++ \cite{liao2018textboxes++} and RRD \cite{liao2018RRD} places 5+ and 10+ anchors with different aspect ratios for short text and long text detection, resulting in a large amount of computational redundancy and long inference time. Anchor clustering is performed in SD \cite{xiao2020SD} to better match the aspect ratios. However, the computed anchors depend on specific datasets. Different from these methods of designing complicated anchors, we turn to the perspective of label assignment since the core reason is few or even no positive samples matching to text instances with extreme aspect ratios.


In this paper, an accurate text detector named MAYOR is proposed to solve these two problems. First, we introduce the learning confusion issue in the mask branch which commonly exists in Mask R-CNN based frameworks. Two ways are proposed to improve the mask branch: (1) We propose to use an MLP decoder instead of the ``deconv-conv'' decoder in the mask head, which alleviates the issue and promotes robustness significantly. (2) We propose instance-aware mask learning (IAML) in which the mask head learns to predict the shape of the whole instance rather than classify each pixel to text or non-text. As shown in Fig. \ref{fig:illus} (b), the quality of the predicted masks is largely promoted with the two proposed techniques. In addition, to meet the demand for large variations of scale and aspect ratio in RPN, we propose a two-step label assignment, namely adaptive label assignment (ALA), in which more positive samples are involved in the pre-assignment process and then $k$ high-quality samples for each ground-truth are selected as the final positive samples according to the matching quality.

The contributions of this work are summarized as follows:
\begin{itemize}
    \item We analyze the performance degradation when Mask R-CNN based frameworks meet dense text detection and argue that the failure originates from the learning confusion issue in the mask head.
    \item A two-layer MLP decoder is proposed to replace the ``deconv-conv'' decoder in the standard mask head, which alleviates the learning confusion issue and promotes robustness significantly.
    \item We propose instance-aware mask learning in which the mask head learns to predict the shape of the whole instance rather than classify each pixel to text or non-text in the pixel-aligned mask learning.  
    \item An adaptive label assignment is proposed in RPN, which brings robust hyper-parameter selection, simple anchor setting, as well as better performance.
    \item Experiments on five public datasets including DAST1500, MSRA-TD500, ICDAR2015, CTW1500, and Total-Text demonstrate the effectiveness of the proposed method. The experimental results also show that the proposed method does not rely on extra pretraining datasets and runs fast in inference. Results on the general instance segmentation task illustrate the generalization of the proposed method.
\end{itemize}


\begin{figure*}[t]
\begin{center}
    \includegraphics[width=0.82\linewidth]{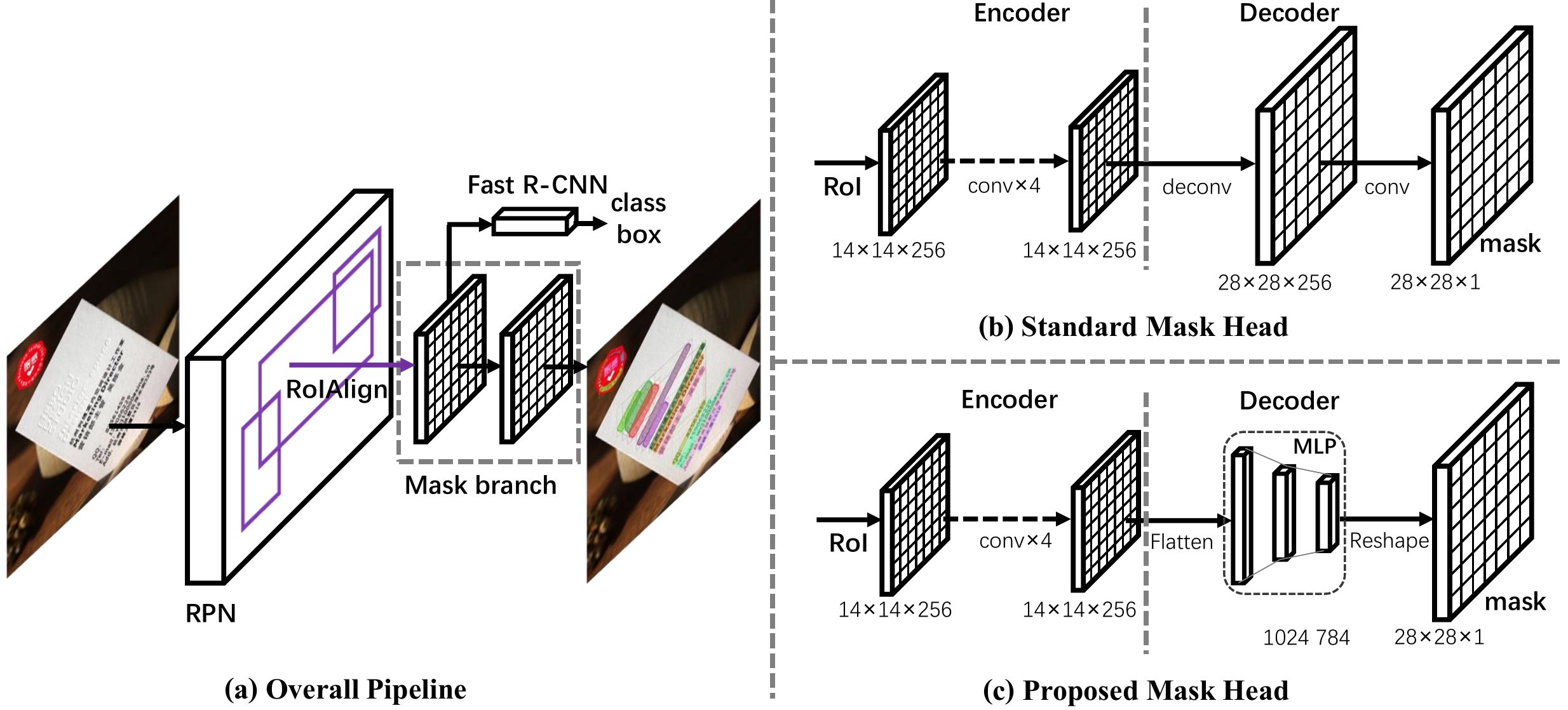}
\end{center}
\vspace{-10px}
    \caption{(a) The overall pipeline of MAYOR. (b) and (c) are the architectures of the standard mask head in Mask R-CNN and the proposed mask head.}
\label{fig:pipeline}
\end{figure*}

\section{Related Work}
According to different perspectives for modeling scene text, scene text detection can be roughly divided into bottom-up and top-down methods. Compared with general objects, the text objects are typically long which require a larger receptive field.

\textbf{Bottom-Up Methods} alleviate the demand by detecting local boxes or pixels and then grouping them into text instances. Segmentation-based methods \cite{deng2018pixellink,wang2019PSENet,tian2019SAE,wang2019PAN,xu2019textfield,liao2020db,wu2017self} predict text or non-text segmentation and other attributes (e.g. similarity embeddings \cite{tian2019SAE,wang2019PAN}, shrink kernels \cite{wang2019PSENet,wang2019PAN}) to group pixels into different instances. Though flexible in the representation of text instances, these methods are sensitive to text-like noise and rely on post-processing. Component-based methods \cite{tian2016CTPN,shi2017SegLink,tang2019seglink++,zhang2020DRRG,ma2019relationtext,baek2019CRAFT,long2018textsnake,feng2019textdragon,zhou2020crnet,cyd1} predict local units and linkages between the components. Lyu et al. \cite{lyu2018corner} propose to first localize corners of text bounding boxes then group the corners into different instances with a position-aware segmentation score. Bottom-up methods could localize local units accurately. However, due to the lack of holistically instance-level supervision, the methods suffer from accumulated errors.

\textbf{Top-Down Methods} directly perform instance-aware prediction, typically comprising one or several stages in a coarse-to-fine manner. These methods adopt global modeling with general object detection frameworks in which multiple detections are generated and Non-Maximum Suppression (NMS) is used to suppress the redundant detections. In regression-based methods, geometry of text is directly predicted from convolutional features \cite{zhou2017east,he2017deepregression,he2017SSTD,liao2018textboxes++,liao2018RRD,wang2018ITN,Liu2020ABCNet,Wang2020textray,wang2019dsrn,liu2017DMPNet,wang2019SAST,qxg1,cyd2} or RoI features \cite{ma2018RRPN,liu2019curved,wang2019ATRR,zhang2018FEN}, and then used to decode to produce the predicted results based on given reference points or boxes. In instance segmentation based methods, typically, Mask R-CNN based methods \cite{lyu2018masktextspotter,xie2019SPCNet,Wang2020contournet,xiao2020SD,liu2019CSE,qxg2}, an extra branch is added to a detection framework. The results are achieved via instance segmentation, getting rids of learning target confusion problem \cite{liu2019BDN,xu2019GNNets} which exists in regression-based methods.
LOMO \cite{zhang2019lomo} is similar to Mask R-CNN on the whole framework, but differs on quadrilateral proposals and a shape expression module to achieve accurate instance-level bounding boxes.
These kinds of methods usually consist of multiple stages, the final instance segmentation results rely heavily on the quality of the detected bounding boxes. 

\textbf{Hybrid Methods.} In addition to the above two types of methods, another trend is to collaborate the bottom-up methods with the top-down frameworks, which can benefit from two modeling perspectives. 
Liao et al. \cite{liao2020masktextspotterv3} propose a segmentation proposal network to replace the original RPN in \cite{liao2019masktextspotterv2}, which make the network robust for text spotting. However, it also suffers from the shortcomings of the segmentation framework. The localization errors from the kernel segmentation degrade the accuracy of the bounding boxes and the discriminant ability of the classifier in the fast R-CNN branch.

\textbf{Detection in Crowded Scenes.} There may be multiple instances falling into a bounding box in crowded scenes, making it difficult to distinguish different instances. Mask R-CNN settles this by class-aware mask prediction. Since the number of dense objects of the same category is relatively rare in typical object detection benchmarks like COCO, this issue does not catch extensive attention of the object detection community. However, it is quite noteworthy in dense text detection which only has a single class. Though difficulty also exists in localizing highly overlapped boxes \cite{chu2020detection}, we find that with sufficient rotation data augmentation, text instances can be well localized with axis-aligned bounding boxes. The focus of this work is learning masks and generating proposals accurately.

In this work, from a top-down perspective, we delve into mask learning and point out the learning confusion issue in the mask head when detecting dense texts. Benefiting from the proposed MLP mask decoder, IAML, and ALA in RPN, our methods effectively handle the learning confusion issue and large aspect ratio variance problem, achieving more accurate mask predictions compared with previous methods. 

\vspace{-5px}
\section{Proposed Method}
In this section, we describe the proposed method in detail. First, we revisit Mask R-CNN for dense scene text detection and introduce the learning confusion issue in pixel-aligned mask learning. An MLP mask decoder is proposed to alleviate it. Next, the concept of instance-aware learning is elaborated. Then we describe the adaptive label assignment. Finally, the optimization is presented.
\vspace{-5px}
\subsection{Scene Text Detection with Mask R-CNN}
\subsubsection{Revisiting Pixel-Aligned Mask Learning}
As shown in Fig. \ref{fig:pipeline} (a), the proposed method follows the overall framework of Mask R-CNN and consists of four modules: a ResNet50-FPN backbone \cite{lin2017FPN} for feature extraction, an RPN module for proposal generation, a fast R-CNN module for refining proposals, and a mask branch for accurate detection. The architecture of the standard mask branch is illustrated in Fig. \ref{fig:pipeline} (b), which can be viewed as an encoder-decoder structure: four convolutions for feature encoding and a deconvolution layer with a convolution predictor for mask decoding. 

Given proposals generated by RPN and multi-scale features generated by FPN, the RoI features are obtained by the RoIAlign operation \cite{he2017maskrcnn}. The mask predictions are produced via the mask branch. In training, binary mask labels are first cropped based on the proposals and then resized to the shape of mask predictions (e.g., $28 \times 28$) to produce the learning targets. In standard Mask R-CNN, the mask predictions are supervised by the corresponding targets in a pixel-aligned manner as shown in Fig. \ref{fig:iaml}. Though the whole modeling follows a top-down perspective, the mask branch is trained to distinguish each pixel belong to text or non-text locally. During inference, the predicted masks are resized according to the predicted boxes and then pasted to the original image.

\begin{figure}[!htb]
\begin{center}
    \includegraphics[width=0.8\linewidth]{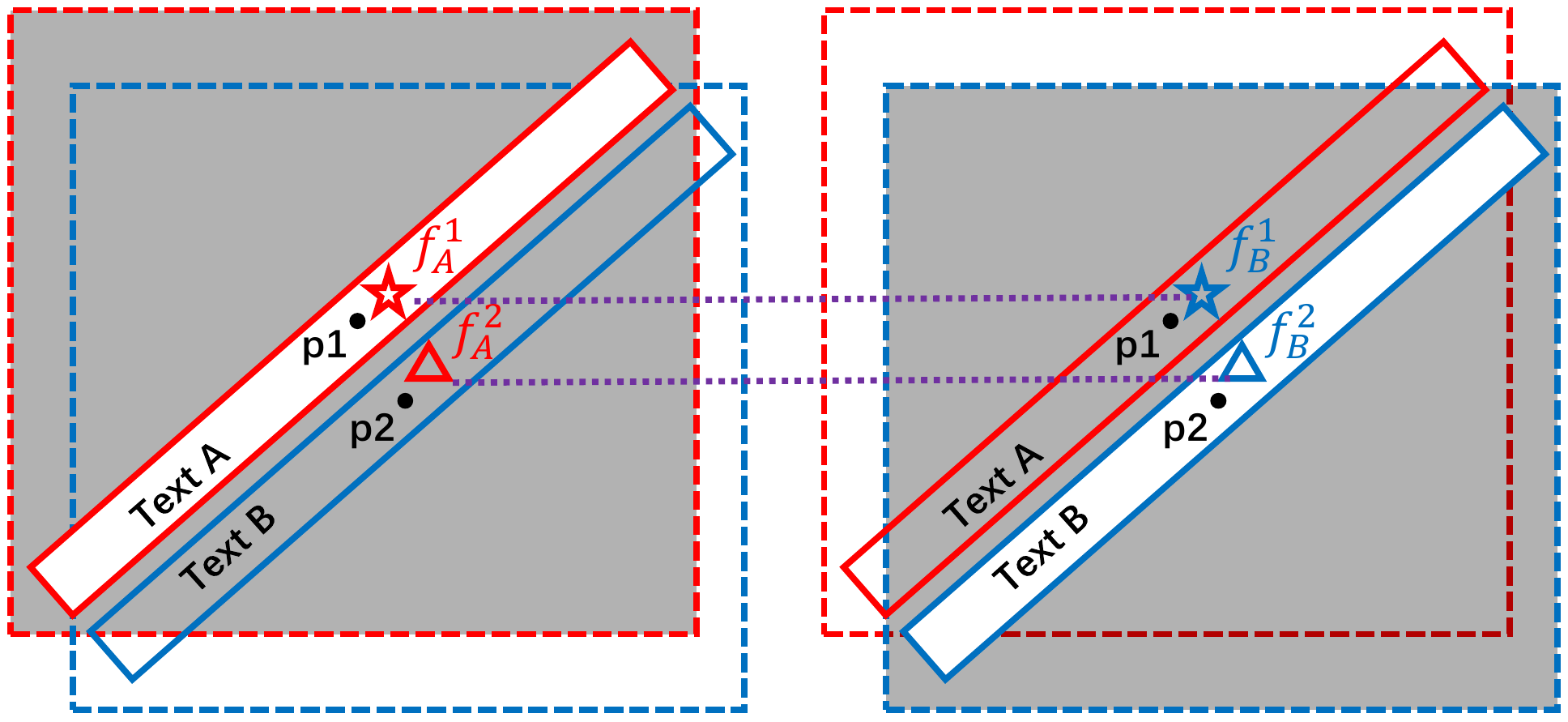}
\end{center}
    \caption{Illustration of the learning confusion issue in mask head. The rotated rectangles and dotted boxes denote different instances and the corresponding proposals generated by RPN. $f_A^1$ and $f_B^1$ denote the closest feature vector to reference point p1 in the RoI feature $F_A$ and $F_B$. $f_A^2$ and $f_B^2$ denote the closest feature vector to reference point p2 in the RoI feature $F_A$ and $F_B$. Spatial closed feature pairs ($f_A^1, f_B^1$), ($f_A^2, f_B^2$) are likely to have very similar features but associated with opposite labels.} 
\label{fig:confusion}
\end{figure}

\vspace{-15px}
\subsubsection{Learning Confusion When Mask R-CNN Meets Dense Texts}
Despite powerful representation for arbitrary-shaped text, Mask R-CNN can not handle dense texts well.
We find that the performance degradation is caused by the learning confusion issue in mask learning.
As illustrated in Fig. \ref{fig:confusion}, given two close text instances A and B, we denote the RoI features encoded by the mask encoder for the two instances as $F_A, F_B \in \mathbb{R}^{N \times N \times C}$. $N$ and $C$ are the dimensions of the spatial output and channel respectively. The mask decoder requires to classify each feature vector $f_A, f_B \in \mathbb{R}^C$ in $F_A, F_B$ to text or non-text. We introduce two reference points p1 and p2 that fall into text A and B respectively. Given p1 and p2, $f_A^1$ and $f_A^2$ denote the nearest feature vectors in $F_A$. Analogously, $f_B^1$ and $f_B^2$ denote the nearest feature vectors to the two points in $F_B$. Since the spatial positions of feature pairs ($f_A^1, f_B^1$) and ($f_A^2, f_B^2$) are quite close, they are likely to have very similar features. When training with $F_A$, $f_A^1$ and $f_A^2$ are taken as positive and negative respectively. When training with $F_B$, $f_B^1$ and $f_B^2$ are taken as negative and positive respectively. The labels are opposite when training with the pairs, resulting in the learning confusion issue and degrading the overall performance.

\subsubsection{MLP Mask Decoder}

\begin{figure}[!htb]
\begin{center}
    \includegraphics[width=0.8\linewidth]{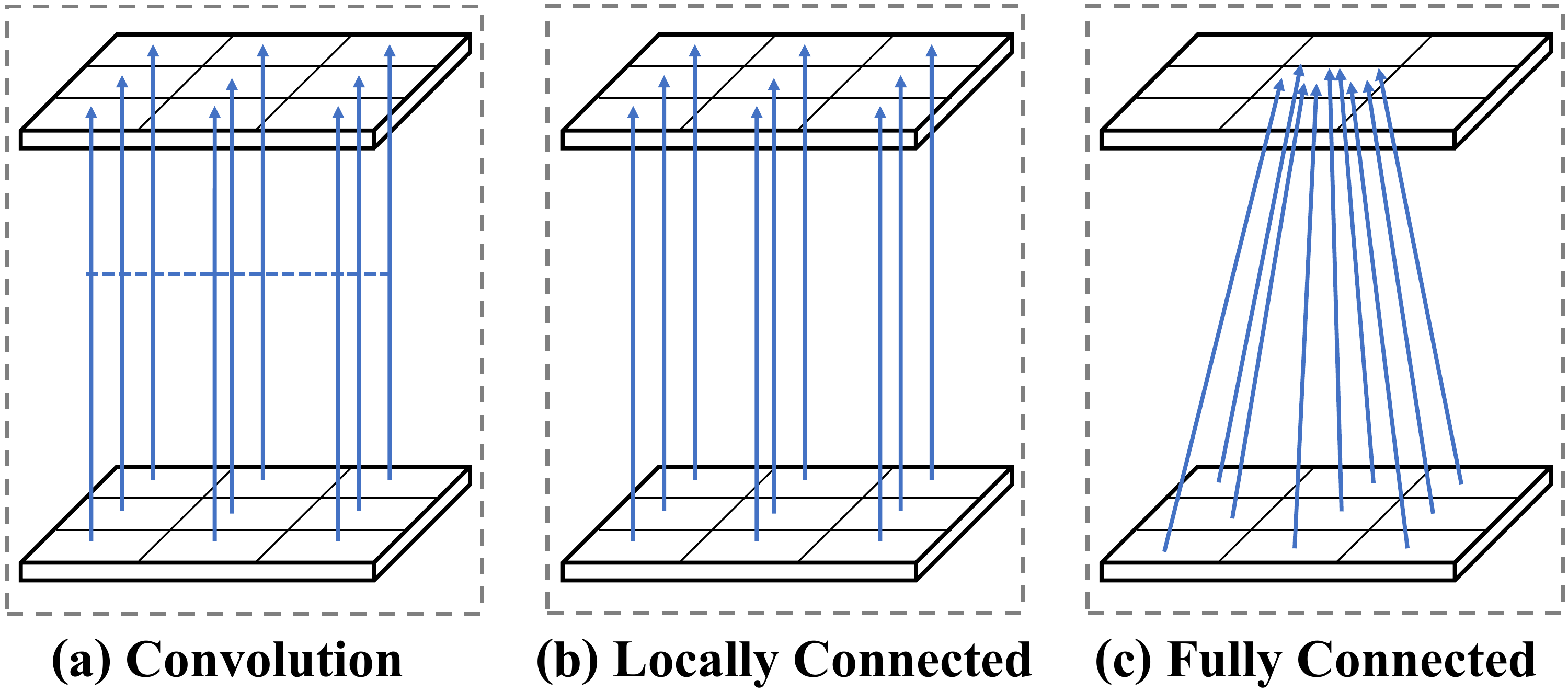}
\end{center}
    \caption{Illustration of different mask predictors. The solid lines denote the data flow. The dotted lines represent sharing weights.}
\label{fig:predictor_type}
\end{figure}

As a basic component in deep neural networks, convolution is popular and widely used due to the benefits of weight sharing and local connectivity. However, we find it harmful to use a convolution classifier for pixel-wise binary classification when detecting dense texts. The learning confusion issue, as mentioned in the last subsection, is magnified by the convolution classifier. The weight sharing makes the weights update more frequently and reduces the discrimination performance of the classifier. To alleviate the learning confusion issue, we design two variants, called locally connected (LC) predictor and fully connected (FC) predictor. Both of them discard weight sharing, while FC predictor makes use of global context information. The predictors of different types are illustrated in Fig. \ref{fig:predictor_type}. 

As shown in Tab. \ref{tab:predictor_type}, without weight sharing, the composed mask decoder with LC predictor (deconv-LC) outperforms the ``deconv-conv'' decoder by 2.8\% F-measure. When we further replace the LC predictor with the FC predictor (deconv-FC) in which more contexts are used to predict the mask, another 0.9\% F-measure increase is obtained. Compared with using more contexts, the discarding of weight sharing brings more obvious improvement with fewer parameters. The reason may be that the RoI features have already obtained a certain extent of receptive field with the mask encoder. The FC predictor also brings in more parameters and computations. In practice, we replace the deconvolution layer with a fully connected with 1024 dimensions (FC-FC) for speed/accuracy trade-off and fair comparison. The computation overhead of the composed two-layer MLP decoder (5.3 GFLOPS) is comparable with the ``deconv-conv'' decoder (5.2 GFLOPS) considering 100 proposals. The head architecture is shown in Fig. \ref{fig:pipeline} (c).

\vspace{-10px}
\begin{table}[!htb]
\small
\caption{Performance with different decoders on DAST1500. Where ``R'', ``P'', ``F'' mean recall, precision, and F-measure respectively. ``deconv'', ``conv'', ``LC'' and ``FC'' are short for deconvolution, convolution, locally connected, and fully connected.}
\vspace{-10px}
\label{tab:predictor_type}
\begin{center}
\begin{tabular}{|c|c|c|c|}
\hline
{\bf Decoder} & {\bf R} & {\bf P} & {\bf F} \\
\hline
deconv-conv & 79.8 & 86.7 & 83.1 \\
deconv-LC & 84.0 & 87.9 & 85.9 \\
deconv-FC & 84.6 & {\bf 89.1} & {\bf 86.8} \\
FC-FC & {\bf 85.5} & 87.8 & 86.6 \\
\hline
\end{tabular}
\end{center}
\end{table}
\vspace{-10px}

\subsection{Instance-Aware Mask Learning}

With the proposed MLP mask decoder, the robustness of the mask head is significantly improved. The separated weights and more contexts with the whole RoI features can alleviate the learning confusion issue.
However, the issue still exists in the standard pixel-aligned mask learning manner. 

\begin{figure}[!htb]
\begin{center}
    \includegraphics[width=0.8\linewidth]{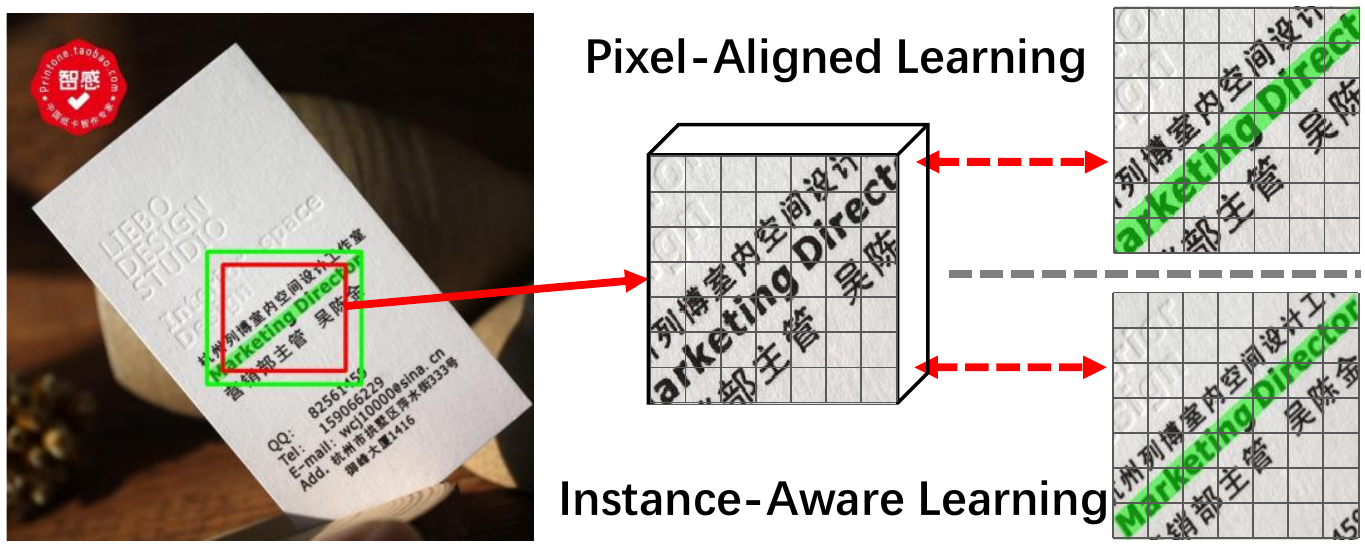}
\end{center}
\vspace{-5px}
    \caption{Illustration of pixel-aligned mask learning and instance-aware mask learning. The red and green boxes denote the RoI and the corresponding ground-truth box respectively.}
\label{fig:iaml}
\end{figure}

To eliminate the issue, we alternatively propose instance-aware mask learning in which the mask branch learns a global instance-aware mask. Instead of classifying pixels into foreground and background in pixel-aligned mask learning, the mask head learns a global representation of how the mask is distributed in the normalized subdivision grids as illustrated in Fig. \ref{fig:iaml}. Compared with the standard pixel-aligned mask learning, the IAML is more like the bounding box regression task, in which a more detailed global shape mask is learned as the coarse bounding box presentation is learned in fast R-CNN branch. We also use the MLP decoder architecture in IMAL since the IAML naturally requires global modeling. In IAML, the learning targets are independent of the proposal boxes and are identical if associated with the same ground-truths.

\vspace{-15px}
\vspace{-15px}



\subsection{Adaptive Label Assignment in Region Proposal Network}
Let us take a brief look at how label assignment is conducted in standard RPN. Given an input image $M$, the ground-truth annotations are denoted as $G$, where a ground-truth box $g_i\in G$
is made up of a class label $g_{i}^{obj}$ and a location $g_{i}^{loc}$. In RPN, $a_j \in A$ stands for an anchor box. Intersection-over-Union (IoU) is used as the matching quality. 
During training, $a_j$ is assigned to GT $g_i$ if $IoU(a_j, g_i^{loc}) > 0.7$, while $a_j$ is defined as negative if $\forall g_i \in G, IoU(a_j, g_i^{loc}) < 0.3$. Anchors which are neither positive nor negative are ignored at the training step. Due to large variations in aspect ratio for scene text, hand-crafted anchor setting is required, and is hard to match texts with extreme aspect ratios, resulting in no or too few positive anchors for these instances and degeneration on performance.

The proposed adaptive label assignment consists of two steps: a pre-assignment step in which abundant samples are likely to be taken as positives and an assignment step in which high-quality samples are further selected as the final positives among the candidates produced in pre-assignment.

\textbf{Label Pre-Assignment.}
The pre-assignment process is the same as the label assignment in standard RPN except that the IoU thresholds for the positives and the negatives are both set to 0. The pre-assignment eliminates the hyper-parameter setting for the thresholds and enables more samples to participate in training. The positive candidates for $g_i\in G$ after this process are denoted as $C^i$.

\textbf{Label Assignment.}
In this process, we need to select high-quality samples from the positive candidates. Inspired by FreeAnchor \cite{Zhang2019FreeAnchor}, we propose to use losses as the measurement of matching quality. 
For each ground-truth $g_i \in G$, we calculate the losses with candidates $c_j \in C^i$ to get {\{$L_{rpn}(g_i, c_j), j=1,...,|C^i|$\}} and pick candidates with the top $k$ smallest values as the final positive samples. Other samples are taken as negatives. As shown in Tab. \ref{tab:topk} and Tab. \ref{tab:ar}, the proposed adaptive label assignment can simplify anchor setting.

\subsection{Optimization}
The loss function $L$ is defined as bellow:
\begin{equation}
L = L_{rpn} + \lambda_{1}L_{rcnn} + \lambda_{2}L_{mask}
\end{equation}
where $L_{rpn}$, $L_{rcnn}$, and $L_{mask}$ are the loss functions defined in RPN,
Fast R-CNN, and Mask R-CNN which are identical as these in \cite{ren2015fasterrcnn,girshick2015fastrcnn,he2017maskrcnn}.
In this work, the $\lambda_{1}$ and $\lambda_{2}$ are empirically set to 1.0.

The learning targets in the pixel-aligned mask learning are identical to those in Mask R-CNN. In IAML, the learning targets are generated with the ground-truth bounding boxes instead of the proposal boxes.

\section{Experiments}
\subsection{Datasets}

\noindent\textbf{DAST1500} \cite{tang2019seglink++} is a dense and arbitrary-shaped text detection dataset, which collects commodity images with a detailed description of the commodities on small wrinkled packages from the Internet. It contains 1038 training images and 500 testing images. Polygon annotations are given at the text line level.

\noindent\textbf{RotDAST} is generated from the DAST1500 dataset \cite{tang2019seglink++}. To test rotation robustness, the dataset is created by rotating the images and annotations in the test set of the DAST1500 with some specific angles, including $0^{\circ}$, $15^{\circ}$, $30^{\circ}$, $45^{\circ}$, $60^{\circ}$, $75^{\circ}$, and $90^{\circ}$. The evaluation protocol is the same as that in DAST1500. 

\noindent\textbf{MSRA-TD500} \cite{yao2012detecting} is a multilingual dataset focusing on oriented text lines. Large variations of text scale and orientation are presented in this dataset. It consists of 300 training images and 200 testing images. Since the training images are too few, we follow the common practice of previous works \cite{zhou2017east,long2018textsnake,wang2019PAN} to add 400 more images from HUST-TR400 \cite{yao2014unified} to the training data.

\noindent\textbf{ICDAR2015 (IC15)} \cite{karatzas2015icdar2015} is a multi-oriented text detection dataset only for English, which includes 1000 training images and 500 testing images. The text regions are annotated with quadrilaterals.

\noindent\textbf{CTW1500} \cite{liu2019curved} contains 1000 training images and 500 testing images.
There are 10751 text instances in total, where 3530 are curved texts and each image has at least one curved text. In this dataset, text instances are annotated with 14-point polygons at the text line level.

\noindent\textbf{Total-Text (TT)} \cite{ch2017total} has 1255 training images and 300 testing images, which contains curved texts, as well as horizontal and multi-oriented texts. Each text is labeled as a polygon in word-level.
\vspace{-10px}

\subsection{Implementation Details}
The model is initialized with ImageNet and trained using SGD optimizer with the learning rate that starts from 0.00125. All the experiments are performed on a GeForce GTX 2080 Ti. The batch size is set to 1. 90k iterations are in total for all datasets. The learning rate decays by 0.1 at 60k and 80k iterations. We adopt random rotation, random crop, random scale as data augmentation in training. Follow the common practice \cite{xie2019SPCNet,liu2019BDN,Wang2020contournet}, the aspect ratios of anchors are set to \{0.25, 0.5, 1.0, 2.0, 4.0\} in the Mask R-CNN baseline.

During inference, predicted masks are converted to polygons by finding connected components. When multiple components case arises, the one with the largest area is picked as the result. Single-scale testing is used for fair comparison and an additional polygonal NMS is adopted to suppress redundant detections. The short side of input images is resized to 800 if not specified.
\vspace{-10px}

\subsection{Ablation Study}
We conduct several ablation studies on DAST1500 to verify the effectiveness of the proposed MLP mask decoder, IAML, and ALA in RPN.

\textbf{MLP Mask Decoder.} To evaluate the effectiveness of the proposed MLP mask decoder, we conduct experiments on DAST1500. As shown in Tab. \ref{tab:dast}, with the proposed MLP mask decoder, a pure increase of 3.5\% is achieved, illustrating the benefits of the unshared weight setting and using more context information. The use of the MLP mask decoder alleviates the learning confusion issue in the mask head and promotes robustness in dense text detection significantly. A decrease is also observed when using IAML. The reason behind this phenomenon is quite simple. Since the IAML is decoupled with the proposal box in learning, it relies more on the accuracy of the detected bounding box. As a contrast, in pixel-aligned learning, the variation of proposal boxes serves as a data augmentation in training, which makes it more robust on unseen images from the testing set.

\begin{table}[!htb]
\small
\caption{Detection results on the DAST1500 dataset. * indicates the results from \cite{tang2019seglink++}. $\dagger$ denotes pretrained with SynthText \cite{gupta2016synthetic}. ``MRCNN'', ``ALA'', and ``MMD'' denote Mask R-CNN, adaptive label assignment in RPN, and the proposed MLP mask decoder.}
\vspace{-5px}
\label{tab:dast}
\begin{center}
\begin{tabular} {|c|c|c|c|}
\hline
{\bf Method} & {\bf R} & {\bf P} & {\bf F} \\
\hline
TextBoxes*++ \cite{liao2018textboxes++} & 40.9 & 67.3 & 50.9 \\
RRD* \cite{liao2018RRD} & 43.8 & 67.2 & 53.0 \\
EAST* \cite{zhou2017east} &  55.7 & 70.0 & 62.0 \\
SegLink* \cite{shi2017SegLink} & 64.7 & 66.0 & 65.3 \\
CTD+TLOC* \cite{liu2019curved} & 60.8 & 73.8 & 66.6 \\
PixelLink* \cite{deng2018pixellink} & 75.0 & 74.5 & 74.7 \\
ICG$\dagger$ \cite{tang2019seglink++} & 79.2 & 79.6 & 79.4 \\
ReLaText$\dagger$ \cite{ma2021relatext} & 82.9 & {\bf 89.0} & 85.8 \\
\hline\hline
{\bf MRCNN}  & 79.0 & 86.3 & 82.5 \\
{\bf MRCNN + ALA} & 79.8 & 86.7 & 83.1\\
{\bf MRCNN + ALA + MMD (MAYOR)} & {\bf 85.5} & 87.8 & {\bf 86.6} \\
\hline
{\bf MAYOR w IAML} & 83.4 & 87.3 & 85.3 \\
\hline
\end{tabular}
\end{center}
\end{table}
\vspace{-15px}

\begin{table}[!htb]
\small
\caption{Detection results on RotDAST when using ground-truths instead of predicted bounding boxes.}
\vspace{-5px}
\label{tab:rotdast_0_45}
\begin{center}
\setlength{\tabcolsep}{1.8mm}
\begin{tabular}{|c|c|c|c|c|c|c|}
\hline
\multirow{2}{*}{\bf Method} &
   \multicolumn{3}{|c|}{\bf RotDAST ($0^{\circ}$)} & \multicolumn{3}{|c|}{\bf RotDAST ($45 ^{\circ}$)} \\
   \cline{2-7}
   & {\bf R} & {\bf P} & {\bf F} & {\bf R} & {\bf P} & {\bf F} \\
\hline
MRCNN & 90.8 & 96.4 & 93.5 & 57.2 & 64.8 & 60.7 \\
MAYOR & 98.2 & 99.0 & 98.6 & 82.9 & 88.2 & 85.5 \\
MAYOR (IAML) & {\bf 99.2} & {\bf 99.4} & {\bf 99.3} & {\bf 96.5} & {\bf 96.9} & {\bf 96.7} \\
\hline
\end{tabular}
\end{center}
\end{table}

\begin{figure}[!htb]
\begin{center}
    \includegraphics[width=0.9\linewidth]{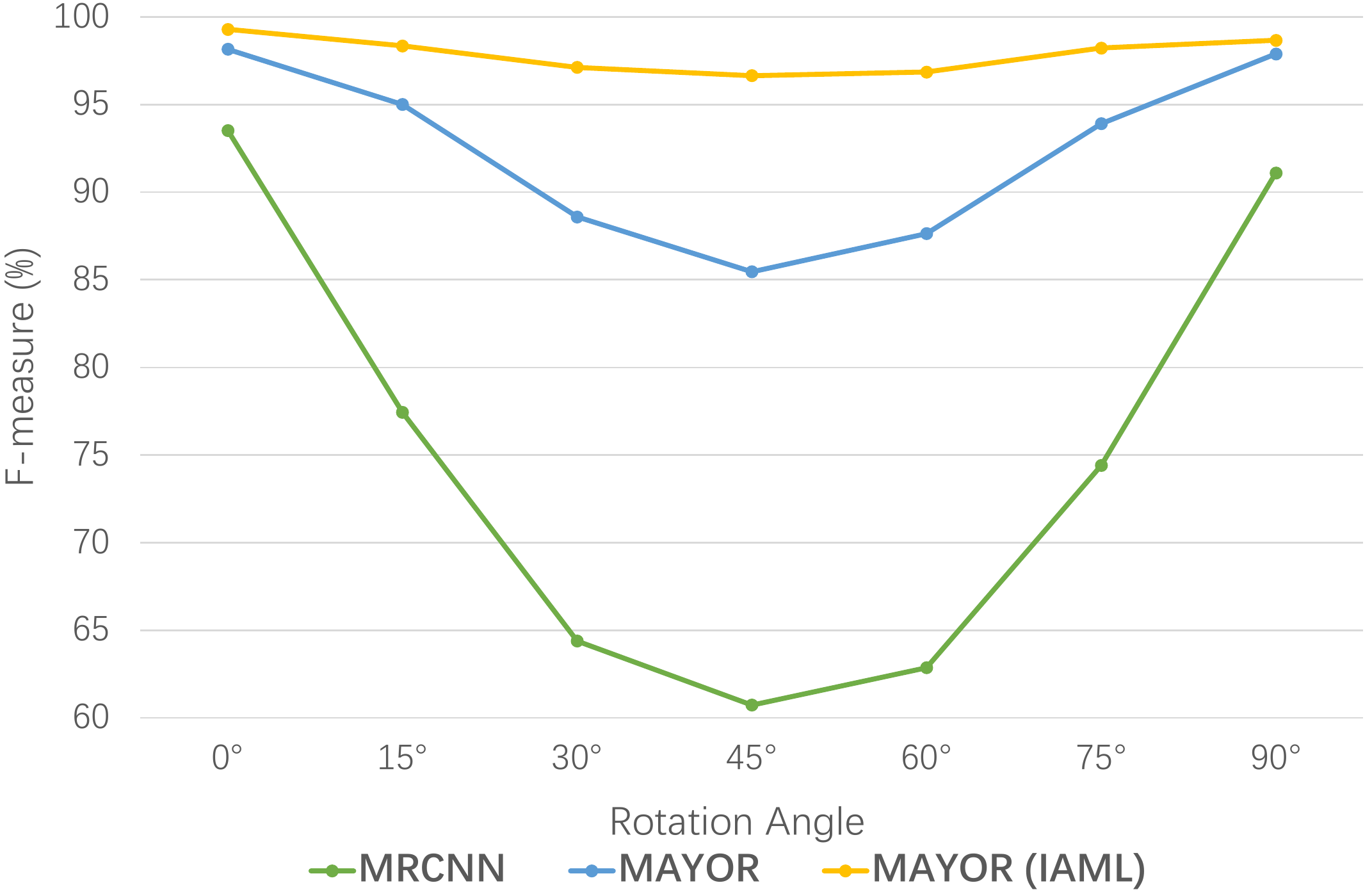}
\end{center}
    \caption{Detection results on RotDAST with different rotation angles when testing with ground-truth bounding boxes.}
    \vspace{-5px}
\label{fig:rotdast_line}
\end{figure}

\textbf{Instance-Aware Mask Learning.} To better illustrate the effectiveness of the proposed IAML which is superior in distinguishing different instances with accurate mask predictions, we further perform experiments on RotDAST. Specifically, we replace the predicted bounding boxes with the corresponding ground-truth boxes to clearly eliminate the impact from bounding box localization. As shown in Tab. \ref{tab:rotdast_0_45}, the MLP mask decoder promotes the Mask R-CNN baseline by 5.1\% in F-measure. With IAML, another 0.7\% increment is obtained. On a more challenging RotDAST with 45 degree, the gaps are further enlarged to 24.8\% and 11.2\% in F-measure. The results demonstrate that the proposed IAML can well distinguish dense text instances. We list results from different degrees to show the robustness of different methods. As shown in Fig. \ref{fig:rotdast_line}, the proposed method with IAML is quite robust with different rotation degrees. As a contrast, the performance of Mask R-CNN drops rapidly when the rotation degree is close to 45. 

\textbf{ALA in RPN.} We first study the robustness of hyper-parameter $k$. As shown in Tab. \ref{tab:topk}, the $k$ value is quite robust across different values. We take $k=5$ as the default setting. Then we perform experiments with anchors of different aspect ratios. The use of ALA enables the number of aspect ratios to be reduced to one as shown in Tab. \ref{tab:ar}. The two observations make the design of anchors more robust and flexible. Furthermore, a study is also conducted with different types of losses as shown in Tab. \ref{tab:loss_type}. Combining with losses of localization and objectness achieves the best performance. Surprisingly, selecting positives with only objectness loss can also achieve an F-measure of 84.4\%. We find that it may come from the strong prior that all positive candidates must have an $IoU > 0$ with a certain ground-truth, which makes the samples quite reliable.

\begin{table}[!htb]
\small
\caption{Detection results on DAST1500 with different values of k.}
\vspace{-5px}
\label{tab:topk}
\begin{center}
\begin{tabular}{|c|ccccccc|}
\hline
{\bf k} & 3 & 5 & 7 & 9 & 11 &  13 & 15 \\
\hline
{\bf R} & 84.8 & {\bf 85.5} & 84.6 & 84.6 & 85.2 &  84.5 & 85.2 \\
\hline
{\bf P} & 88.2 & 87.8 & {\bf 88.5} & 88.1 & 88.0 &  88.4 & 87.7 \\
\hline
{\bf F} & 86.5 & {\bf 86.6} & 86.5 & 86.3 & {\bf 86.6} &  86.4 & 86.4 \\
\hline
\end{tabular}
\end{center}
\end{table}
\vspace{-15px}

\begin{table}[!htb]
\small
\caption{Detection results on DAST1500 with different aspect ratios of anchors.}
\vspace{-5px}
\label{tab:ar}
\begin{center}
\begin{tabular}{|c|c|c|c|}
\hline
{\bf Aspect Ratio} & {\bf R} &{\bf P} & {\bf F} \\
\hline
\{1.0\} & {\bf 85.5} & 87.8 & {\bf 86.6} \\ 
\{0.5, 1.0, 2.0\} & 83.7 & {\bf 89.0} & 86.3 \\
\{0.25, 0.5, 1.0, 2.0, 4.0\} & 84.8 & 88.4 & 86.5 \\
\hline
\end{tabular}
\end{center}
\end{table}
\vspace{-15px}

\begin{table}[!htb]
\small
\caption{Detection results on DAST1500 with different loss settings in adaptive label assignment in RPN, ``Loc'' and ``Obj'' denote localization and objectness.}
\vspace{-5px}
\label{tab:loss_type}
\begin{center}
\begin{tabular}{|cc|c|c|c|}
\hline
{\bf Loc} & {\bf Obj} & {\bf R} & {\bf P} & {\bf F}  \\
\hline
$\checkmark$ & & 82.8 & {\bf 90.1} & 86.3 \\ 
 & $\checkmark$ & 82.9 & 85.9 & 84.4 \\ 
$\checkmark$ & $\checkmark$ & {\bf 85.5} & 87.8 & {\bf 86.6} \\
\hline
\end{tabular}
\end{center}
\end{table}

\begin{table*}[t]
\small
\caption{The single-scale results on ICDAR2015, CTW1500, and Total-Text. Ext is the short for external data used in training stage. ``ST'' and ``MLT'' denote SynthText \cite{gupta2016synthetic} and ICDAR2017-MLT \cite{nayef2017icdar2017mlt}.}
\vspace{-15px}
\label{tab:sotas}
\begin{center}
\setlength{\tabcolsep}{1.5mm}
\begin{tabular}{|c||c|c|c|c|c||c|c|c|c|c||c|c|c|c|c|}
\hline
\multirow{2}{*}{\bf Method} &
   \multicolumn{5}{|c||}{\bf ICDAR2015} & \multicolumn{5}{|c||}{\bf CTW1500} & \multicolumn{5}{|c|}{\bf Total-Text} \\
   \cline{2-16}
   & {\bf Ext} & {\bf R} & {\bf P} & {\bf F} & {\bf FPS} & {\bf Ext} & {\bf R} & {\bf P} & {\bf F} & {\bf FPS} & {\bf Ext} & {\bf R} & {\bf P} & {\bf F} & {\bf FPS}\\ 
\hline
PSENet \cite{wang2019PSENet} & - & 79.7 & 81.5 & 80.6 & 1.6 & - & 75.6 & 80.6 & 78.0 & 3.9 & - & 75.1 & 81.8 & 78.3 & 3.9 \\
ATRR \cite{wang2019ATRR} & - & 86.0 & 89.2 & 87.6 & - & - & 80.2 & 80.1 & 80.1 & - & - & 76.2 & 80.9 & 78.5 & - \\
PAN \cite{wang2019PAN} & - & 77.8 & 82.9 & 80.3 & {\bf 26.1} & - & 77.7 & 84.6 & 81.0 & {\bf 39.8} & - & 79.4 & 88.0 & 83.5 & {\bf 39.6} \\
ContourNet \cite{Wang2020contournet} & - & 86.1 & 87.6 & 86.9 & 3.5 & - & 84.1 & 83.7 & 83.9 & 4.5 & - & 83.9 & 86.9 & 85.4 & 3.8 \\ 
TextSnake \cite{long2018textsnake} & ST & 80.4 & 84.9 & 82.6 & 1.1 & ST & {\bf 85.3} & 67.9 & 75.6 & - & ST & 74.5 & 82.7 & 78.4 & - \\
TextField \cite{xu2019textfield} & ST & 83.9 & 84.3 & 84.1 & 1.8 & ST & 79.8 & 83.0 & 81.4 & 6.0 & ST & 79.9 &81.2 & 80.6 & 6.0 \\ 
LOMO \cite{zhang2019lomo} & ST & 83.5 & 91.3 & 87.2 & 3.4 & ST & 69.6 & {\bf 89.2} & 78.4 & 4.4 & ST & 75.7 & 88.6 & 81.6 & 4.4 \\
SAE \cite{tian2019SAE} & ST & 85.0 & 88.3 & 86.6 & - & ST & 77.8 & 82.7 & 80.1 & - & - & - & - & - & - \\
MSR \cite{xue2019msr} & ST & 78.4 & 86.6 & 82.3 & 4.3 & ST & 78.3 & 85.0 & 81.5 & 4.3 & ST & 74.8 & 83.8 & 79.0 & 4.3 \\
PAN \cite{wang2019PAN} & ST & 81.9 & 84.0 & 82.9 & {\bf 26.1} & ST & 81.2 & 86.4 & 83.7 & {\bf 39.8} & ST & 81.0 & 89.3 & 85.0 & {\bf 39.6} \\
DB \cite{liao2020db} & ST & 83.2 & 91.8 & 87.3 & 12.0 & ST & 80.2 & 86.9 & 83.4 & 22.0 & ST & 82.5 & 87.1 & 84.7 & 32.0 \\
\hline
SPCNet \cite{xie2019SPCNet} & ST+MLT & 85.8 & 88.7 & 87.2 & - & - & - & - & - & - & ST+MLT & 82.8 & 83.0 & 82.9 & - \\
PSENet \cite{wang2019PSENet} & MLT & 84.5 & 86.9 & 85.7 & 1.6 & MLT & 79.7 & 84.8 & 82.2 & 3.9 & MLT & 78.0 & 84.0 & 80.9 & 3.9 \\
CRAFT \cite{baek2019CRAFT} & ST & 84.3 & 89.8 & 86.9 & - & ST+MLT & 81.1 & 86.0 & 83.5 & - & ST+MLT & 79.9 & 87.6 & 83.6 & - \\
DRRG \cite{zhang2020DRRG} & ST+MLT & 84.7 & 88.5 & 86.6 & - & ST+MLT & 83.0 & 85.9 & 84.5 & - & ST+MLT & 84.9 & 86.5 & 85.7 & - \\
SD \cite{xiao2020SD} & MLT & {\bf 88.4} & 88.7 & 88.6 & - & TT+MLT & 82.3 & 85.8 & 84.0 & - & MLT & 84.7 & 89.2 & 86.9 & - \\
\hline \hline
{\bf MRCNN} & - & 82.0 & 90.1 & 85.9 & 6.7 & - & 81.4 & 87.0 & 84.1 & 19.7 & - & 82.3 & 88.9 & 85.5 & 19.7 \\ 
{\bf MAYOR} & - & 85.2 & 90.5 & 87.8 & 7.0 & - & 82.7 & 88.0 & 85.3 & 19.9 & - & 84.5 & 88.2 & 86.3 & 19.9 \\
{\bf MAYOR (IAML)} & - & 85.5 & 89.7 & 87.6 & 7.0 & - & 81.0 & 89.0 & 84.9 & 19.9 & - & 84.2 &  87.8 & 86.0 & 19.9 \\
\hline
{\bf MAYOR} & MLT & 85.9 & {\bf 92.7} & 89.2 & 7.0 & MLT & 83.6 & 88.7 & {\bf 86.1} & 19.9 & MLT & 85.3 & {\bf 92.9} & {\bf 88.9} & 19.9 \\
{\bf MAYOR (IAML)} & MLT & 87.3 & 91.5 & {\bf 89.3} & 7.0 & MLT & 82.1 & 88.7 & 85.3 & 19.9 & MLT & {\bf 87.1} &  90.7 & {\bf 88.9} & 19.9 \\
\hline
\end{tabular}
\end{center}
\end{table*}

\subsection{Comparison with State-of-the-Art Methods}

We compare our MAYOR with recent state-of-the-art methods on DAST1500, MSRA-TD500, ICDAR2015, CTW1500, and Total-Text to demonstrate its effectiveness for dense and arbitrary-shaped text detection.

\subsubsection{Evaluation on Dense and Arbitrary-Shaped Text Benchmark}
We evaluate the proposed method on DAST1500 to test its performance for dense and arbitrary-shaped texts. As shown in Tab. \ref{tab:dast}, with the help of ALA and MMD, MAYOR achieves a new state-of-the-art result of 85.5\%, 87.8\%, and 86.6\% in recall, precision, and F-measure respectively without external data, and outperforms ICG \cite{tang2019seglink++} by a large margin. Though ReLaText \cite{ma2021relatext} uses a powerful graph convolutional network and additional pretraining data, our method trained with only original annotations outperforms ReLaText by 0.8\% in F-measure. Compared with state-of-the-art methods which mostly utilize flexible bottom-up representations, MAYOR directly works in a top-down manner, which also demonstrates top-down modeling can also perform well on dense texts.

\begin{figure*}[t]
\begin{center}
    \includegraphics[width=0.8\linewidth]{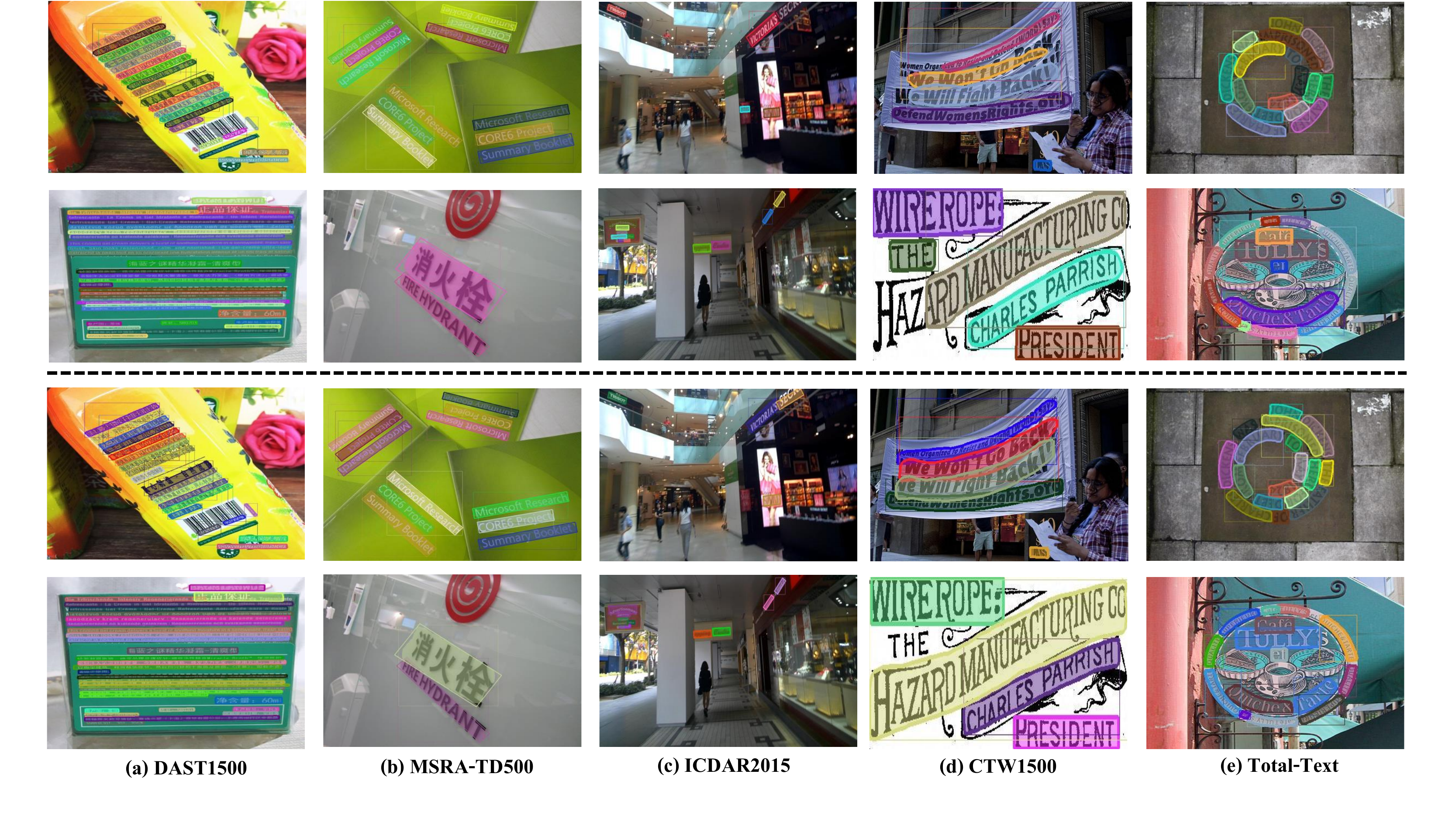}
\end{center}
\vspace{-13px}
    \caption{Qualitative results on DAST1500, MSRA-TD500, ICADR2015, CTW1500, and Total-Text. The first two rows and the last two rows are results from MAYOR and MAYOR (IAML), respectively.}
\label{fig:results}
\end{figure*}

\subsubsection{Evaluation on Multi-Oriented Text Benchmark}
We evaluate our method on MSRA-TD500 and ICDAR2015 to test its performance for multi-oriented texts. As shown in Tab. \ref{tab:td500}, MAYOR achieves 85.2\%, 91.7\%, and 88.3\% on MSRA-TD500 without external data, outperforming existing SOTA methods (e.g. DB \cite{liao2020db}, DRRG \cite{zhang2020DRRG}, ReLaText \cite{ma2021relatext}) by a large margin. It also outperforms MTS v3 \cite{liao2020masktextspotterv3} which uses a segmentation-based method to generated proposals. Compared with MTS v3, MAYOR benefits from global modeling in which proposals are further refined in fast R-CNN branch, alleviating the accumulated localization errors from proposal generation.

Since the scales of texts in ICDAR1205 are generally small, we follow the common practice to use a larger scale input in which the short side to resized to 1440 for testing. On ICDAR2015, MAYOR achieves 87.8\% and 87.6\% (IAML) F-measure without external data as shown in Tab. \ref{tab:sotas}, which outperforms most methods and is only inferior to SD \cite{zhang2020DRRG} that uses ICDAR2017-MLT. To fairly compared with methods that use external data, MLT is also combined as pretraining data. With MLT, MAYOR achieves state-of-the-art results of 89.2\% and 89.3\% (IAML), indicating the effectiveness of detecting multi-oriented texts.

\begin{table}[b]
\small
\caption{The single-scale results on MSRA-TD500. * indicates multi-scale testing. ``RCTW'' denotes RCTW17 \cite{shi2017rctw17}.}
\label{tab:td500}
\begin{center}
\setlength{\tabcolsep}{1.3mm}
\begin{tabular}{|c|c|c|c|c|c|}
\hline
{\bf Method} & {\bf Ext} & {\bf R} & {\bf P} & {\bf F} & {\bf FPS} \\
\hline
EAST \cite{zhou2017east} & - & 67.4 & 87.3 & 76.1 & 13.2 \\
RRPN \cite{ma2018RRPN} & - & 68.0 & 82.0 & 74.0 & - \\
ITN \cite{wang2018ITN} & - & 72.3 & 90.3 & 80.3 & - \\
Border* \cite{xue2018border} & - & 77.4 & 83.0 & 80.1 & - \\
PAN \cite{wang2019PAN} & - & 77.3 & 80.7 & 78.9 & 30.2 \\
ATRR \cite{wang2019ATRR} & - & 82.1 & 85.2 & 83.6 & - \\
SegLink \cite{shi2017SegLink} & ST & 70.0 & 86.0 & 77.0 & 8.9 \\
RRD \cite{liao2018RRD} & ST & 73.0 & 87.0 & 79.0 & 10.0 \\
Corner \cite{lyu2018corner} & ST & 76.2 & 87.6 & 81.5 & 5.7 \\
MCN \cite{liu2018MCN} & ST & 79.0 & 88.0 & 83.0 & - \\
TextSnake \cite{long2018textsnake} & ST & 73.9 & 83.2 & 78.3 & 1.1 \\
SAE \cite{tian2019SAE} & ST & 81.7 & 84.2 & 82.9 & - \\
TextField \cite{xu2019textfield} & ST & 75.9 & 87.4 & 81.3 & - \\  
DB \cite{liao2020db}  & ST & 79.2 & 91.5 & 84.9 & {\bf 32.0} \\
MTS v3 \cite{liao2020masktextspotterv3}& ST & 90.7 & 77.5 & 83.5 & - \\
ReLaText \cite{ma2021relatext} & ST & 83.2 & 90.5 & 86.7 & 8.3 \\
PixelLink \cite{deng2018pixellink} & IC15 & 73.2 & 83.0 & 77.8 & 3.0 \\
SBD \cite{liu2019BDN} & RCTW & 80.5 & 89.6 & 84.8 & 3.2 \\
CRAFT \cite{baek2019CRAFT} & MLT & 78.2 & 88.2 & 82.9 & 8.6 \\
DRRG \cite{zhang2020DRRG} & ST+MLT & 82.3 & 88.1 & 85.1 & - \\
\hline
{\bf MAYOR} & - & {\bf 85.2} & {\bf 91.7} & {\bf 88.3} & 20.5 \\
\hline
\end{tabular}
\end{center}
\end{table}

\subsubsection{Evaluation on Curved Text Benchmark}
To show the performance of our method for curved texts, we compare its performance with the state-of-the-arts on CTW1500 and Total-Text. As shown in Tab. \ref{tab:sotas}, the proposed method is much better than other methods including TextSnake \cite{long2018textsnake}, MSR \cite{xue2019msr}, and DB \cite{liao2020db}, which are designed for curved texts. MAYOR achieves 85.3\% and 84.9\% (IAML) in F-measure on CTW1500 without external data and outperforms recently proposed methods ContourNet \cite{Wang2020contournet}, DRRG \cite{zhang2020DRRG}, and SD \cite{xiao2020SD}. With MLT as pretraining data, it achieves state-of-the-art result of 86.1\% in F-measure. Compared with SD, which is also a Mask R-CNN based framework, the proposed ALA in RPN can exploit more samples for long texts with extreme aspect ratios. And the proposed MMD and IAML are also more robust in detecting long texts. 

On Total-Text, MAYOR achieves 86.3\% and 86.0\% (IAML) in F-measure without external data, outperforming existing methods except for SD which uses ICADAR2017-MLT as the external data. When MLT is also combined in training, a state-of-the-art result of 88.9\% in F-measure is achieved, outperforming SD by 2\%. The performance across different datasets shows the effectiveness and robustness in various scene text cases. Detection results on five datasets are visualized in Fig. \ref{fig:results}.

\vspace{-9px}
\vspace{-10px}
\subsection{General Instance Segmentation on COCO}
\begin{figure}[!htb]
\begin{center}
    \includegraphics[width=1.0\linewidth]{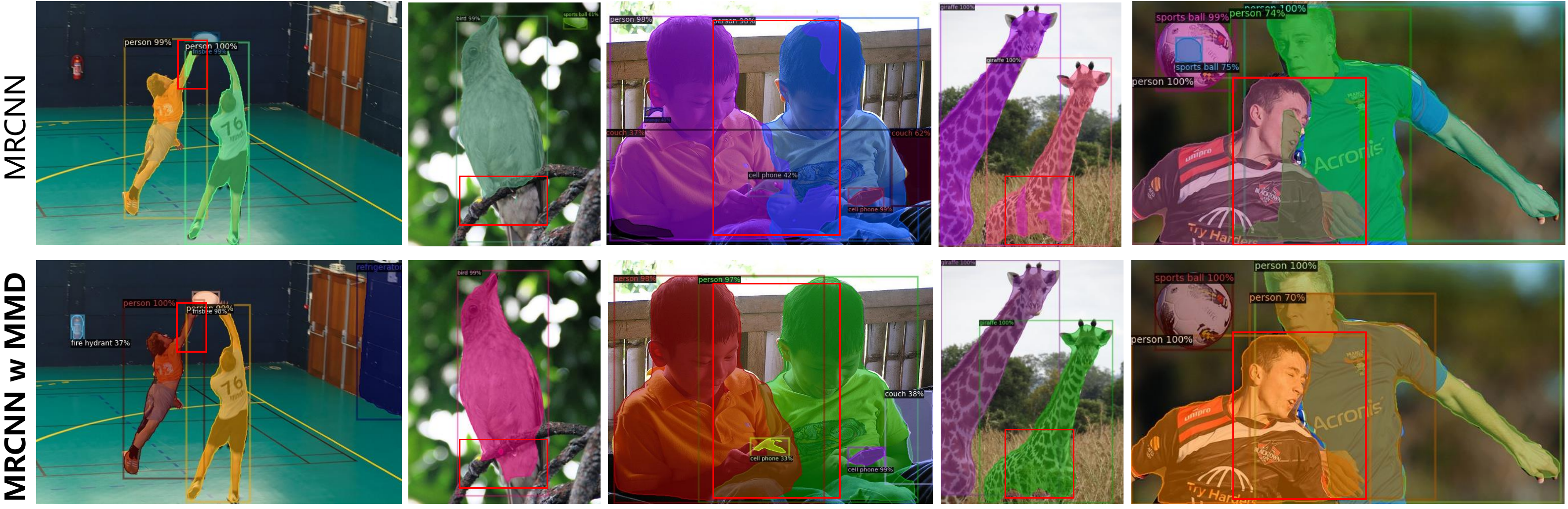}
\end{center}
    \vspace{-13px}
    \caption{Visualization results on COCO.}
    \vspace{-10px}
\label{fig:coco_results}
\end{figure}

The proposed method is general since we consider texts as one type of instance during modeling. When MMD is added to MRCNN, we find an increment from 35.1\% to 35.6\% on mask AP on COCO validation set, which illustrates its effectiveness. Though close instances within the same category are relatively rare on COCO, learning confusion exists and results in fragment prediction. We observe 2\%-5\% AP improvement on snowboard, hot dog, and toothbrush which are also long and thin. As shown in Fig. \ref{fig:coco_results}, the prediction is more compact and complete due to global modeling.
\vspace{-10px}

\section{Conclusion}
In this work, we rethink the limitation of Mask R-CNN based methods on dense scene text detection and point out the learning confusion issue in the mask head. An MLP mask decoder is proposed to alleviate the issue. We alternatively propose instance-aware mask learning to eliminate this issue from a global perspective. The adaptive label assignment is also designed for better matching texts with extreme aspect ratios. With the proposed techniques, our method, namely MAYOR, can better detect dense and arbitrary-shaped text. The performance on five public benchmarks demonstrates the effectiveness and robustness of our approach. We'd like to combine text recognition \cite{qz1,qz2}, self-supervised learning \cite{zyf1,zyf2,ldz1,ldz2,ldz3,lw1}, and knowledge distillation \cite{ydb1,ydb2} to build a robust text reading system.

\begin{acks}
This work is supported by the Open Research Project of the State Key Laboratory of Media Convergence and Communication, Communication University of China, China (No. SKLMCC2020KF004), the Beijing Municipal Science \& Technology Commission (Z1911000\\07119002), the Key Research Program of Frontier Sciences, CAS, Grant NO ZDBS-LY-7024, the National Natural Science Foundation of China (No. 62006221), and CAAI-Huawei MindSpore Open Fund.
\end{acks}

\bibliographystyle{ACM-Reference-Format}
\bibliography{DSTD}

\end{document}